# Efficient Visual Coding: From Retina To V2

**Honghao Shan    Garrison Cottrell**
Computer Science and Engineering
UCSD
La Jolla, CA 92093-0404
shanhonghao@gmail.com, gary@ucsd.edu

**Abstract**

The human visual system has a hierarchical structure consisting of layers of processing, such as the retina, V1, V2, etc. Understanding the functional roles of these visual processing layers would help to integrate the psychophysiological and neurophysiological models into a consistent theory of human vision, and would also provide insights to computer vision research. One classical theory of the early visual pathway hypothesizes that it serves to capture the statistical structure of the visual inputs by efficiently coding the visual information in its outputs. Until recently, most computational models following this theory have focused upon explaining the receptive field properties of one or two visual layers. Recent work in deep networks has eliminated this concern, however, there is till the retinal layer to consider. Here we improve on a previously-described hierarchical model Recursive ICA (RICA) [1] which starts with PCA, followed by a layer of sparse coding or ICA, followed by a component-wise nonlinearity derived from considerations of the variable distributions expected by ICA. This process is then repeated. In this work, we improve on this model by using a new version of sparse PCA (sPCA), which results in biologically-plausible receptive fields for both the sPCA and ICA/sparse coding. When applied to natural image patches, our model learns visual features exhibiting the receptive field properties of retinal ganglion cells/lateral geniculate nucleus (LGN) cells, V1 simple cells, V1 complex cells, and V2 cells. Our work provides predictions for experimental neuroscience studies. For example, our result suggests that a previous neurophysiological study improperly discarded some of their recorded neurons; we predict that their discarded neurons capture the shape contour of objects.

## 1    Introduction

The visual layers that appear early in the human visual pathway, such as retina, V1, and V2, have been of particular interest to theoretical neuroscientists and computer vision researchers. First, these early visual layers capture simpler visual structures and receive less top-down influence than the later visual layers. As a result, experimental neuroscientists have collected detailed descriptions of the early visual layers. Second, these early visual layers are shared components in the visual pathway; they develop and mature in our early life, and then subserve all kinds of visual tasks during the rest of life. Hence they appear to encode certain universal visual features that work as building blocks of visual scenes. It is unlikely that these features are learned in a supervised manner. Hence, it is of particular interest to understand what kinds of visual features they encode, and how these features are learned automatically from natural scenes.

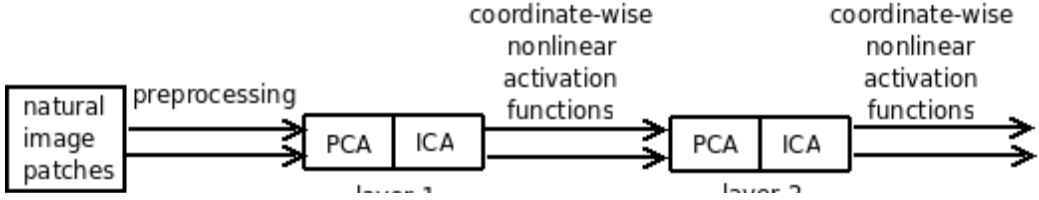

Figure 1. The model framework. Image patches are preprocessed and passed to the first layer. Within the first layer, dimensions of the data are first reduced by the sparse autoencoder (sPCA); then the dimensions are expanded using the sparse coding algorithm (ICA). The layer-1 outputs are transformed to the layer-2 inputs by coordinate-wise nonlinear activation functions.

The efficient coding theory has been popular in explaining the functional roles of the early visual layers. It follows the long-standing hypothesis that the visual perception serves to capture the statistical structure of the visual inputs, so that actions can be taken accordingly to maximize the chances of survival. Attneave suggested that the visual structure could be measured by the redundancy of the visual inputs: how much visual structure we perceive depends on how well we can predict a removed piece of the image by observing its remaining parts. Barlow further hypothesized that the sensory system could capture the statistical structure of its inputs by removing the redundancy in its outputs, because to do so the system must have a complete knowledge of the statistical structure. Linear implementations of the efficient coding theory, such as independent components analysis and sparse coding, have been used to explain the receptive field properties of V1 simple cells. When these algorithms are applied to natural image patches, they learn visual features resembling the edge/bar shaped receptive fields of V1 simple cells.

Here we propose a hierarchical model following the efficient coding principle, as illustrated in Figure 1. This model is a variant on the Recursive ICA model (RICA) [1]. The model has two characteristics. First, this model adopts a layered framework: we apply the efficient coding algorithm to capture the statistical structure of the inputs, and then apply another layer of efficient coding algorithm to capture higher-order statistical structure not captured by the previous layer. Such a layered framework was motivated by the observation that different parts of the brain share similar anatomical structure and hence are likely to work under similar computational principles. Second, within each layer we first apply sparse Principal Components Analysis (SPCA) (see Methods) to reduce the dimensionality of the data, and then use the sparse coding algorithm to expand the dimensionality. This scheme is widely used in computer vision algorithms: we usually apply PCA to reduce dimensionality before applying ICA to capture the statistical structure of the inputs, which helps to reduce noise as well as computational complexity. Moreover, a previous study has shown that SPCA learns the centre-surround shaped filters resembling the receptive fields of retinal ganglion cells and lateral geniculate nucleus cells; applying the sparse coding algorithm afterwards produces the edge/bar-shaped basis functions resembling the receptive fields of V1 simple cells. In this work, we show that applying SPCA and then overcomplete ICA to the result of these first two layers of processing gives V1 complex cell receptive fields followed by receptive fields that explain data recently reported concerning V2, respectively.

One difference between previous work in multi-layer ICA and ours is the application of a component-wise nonlinearity to the outputs of the ICA layer. To allow the second layer to more efficiently capture the statistical structure in the first layer outputs, we derived coordinate-wise nonlinear activation functions to transform the first layer's outputs to the second layer's inputs. The activation functions contain two steps: (1) take the absolute values; (2) make the marginal distribution a Gaussian distribution. The idea here is that, as a generative model, ICA generates the input pixels as a linear combination of many independent variables, which results in a Gaussian distribution. Hence, our transformation "formats" the outputs of ICA for the next layer up.

## 2    Methods

**Image dataset.** We use the sixty-two 1000 by 1280 colour images of the natural scenes taken near Kyoto. We transform them to grey-scale images, down-scale them by half on both sides. We then normalize each image to have zero-mean and unit-variance pixel values.

**Sparse PCA.** We began with consideration of the constraints on retinal ganglion cells. The retina compresses the approximately 100 million photoreceptor responses into a million ganglion cell re-sponses. Hence the first consideration is that we would like the ganglion cells to retain the maximum amount of information about the photoreceptor responses. If we make the simplifying assumption that ganglion cells respond linearly, then the optimal linear compression technique in terms of reconstruction error is principal components analysis (PCA). One can map PCA into a neural network as in Figure 1(a) (18,19). The weight vectors of each hidden unit in this network each correspond to one eigenvector of the covariance matrix of the data. In standard PCA, there is an ordering to the hidden units, such that the first hidden unit has very high response variance and the last hidden unit has practically no variance, which means the first hidden unit is doing orders of magnitude more work than the last one. The second consideration, then, is that we would like to spread the work evenly among the hidden units. Hence we impose a threshold on the average squared output of the hidden units. As we will see from the simulations, in order to preserve the maximum information, the units all hit this threshold, which equalizes the work. The third consideration is that PCA is profligate with connections - every ganglion cell would have non-zero connections to every photoreceptor. Hence we also impose a constraint on the connectivity in the network. In this latter constraint we were inspired by the earlier work of (7). They proposed a model of retinal and early cortical processing based on energy minimization and showed that it could create center-surround shaped receptive fields for grayscale images. However, their system sometimes led to cells with two center-surround fields, and the optimization itself was unstable.

The three principles described above leads to our objective function. Given an input vector $x \in R^L$, the autoencoder seeks the optimal $s^* \in R^M$ and $A \in R^{L \times M}$ such that the following objective function is minimized:

$$E = \left\langle \frac{\|x - As\|_2^2}{2} \right\rangle + \lambda \|A\|_1 \quad \text{subject to: } \langle s_i^2 \rangle \leq 1 \quad \forall i$$

where the angle brackets denote taking average over all the input samples. The first term minimizes the reconstruction error, maximizing the information retained by the encoding. When $\lambda$ is small, and L > M (i.e., the encoding compresses the information), the reconstruction error is well approximated by a term that only involves the correlation matrix: $C = \langle xx' \rangle$, which leads to a fast approximation algorithm. Given that the second term minimizes the connections, and the system is sensitive to correlations, this leads to local receptive fields. The final constraint equalizes the work, and the system pushes up against this limit, so that the average variance is equal to 1.

We set lambda = 0.01 for the first layer, and lambda=0.03 for the second layer. We use gradient descent to find the optimal $s$. We start with $s=A'x$, and update $s$ for 100 steps with an update rate of 0.01.

**Sparse coding.** Given an input vector $x \in R^L$, the sparse coding algorithm seeks the optimal $s^* \in R^M$ that minimizes the reconstruction error as well as a sparsity penalty on the outputs:

$$E = \left\langle \frac{\|x - As\|_2^2}{2} \right\rangle + \lambda f(s)$$

Once the optimal $s$ is inferred, the connection weights $A$ are updated by gradient descent in $E$. At this point, as in previous work [1], a coordinate-wise nonlinear activation function is applied to each filter. Our goal is to find the coordinate-wise activation function $g_i$ for each dimension of $s$ such that $x_i=g_i(|s_i|)$ follows a generalized Gaussian distribution (in this work, we simply assume a Gaussian). Note here, as in previous work, we discard the sign on $s_i$. We use a non-parametric approach for efficiency. In this approach, all the samples $|s_i|$ are sorted in ascending order. For $N$ samples of each coordinate $s_i$, $cdf(|s_i|)$ is approximated by the ratio of its ranking in the list with $N$. Then $x_i = g(|s_i|) = F^{-1}(c\hat{df}(|s_i|))$, where $F$ is the cumulative density function of a normal distribution, will approximately follow the standard normal distribution. Note that since $x_i$ depends only on the rank order of $|s_i|$, the results would be the same if the signs are discarded by squaring the $s_i$'s.

**Model training.** The model is trained layer by layer. For the first layer sparse autoencoder, we randomly initialize A with Gaussian random variables, and update A for 40000 epochs. During each epoch, we randomly select one image and sample 64 16x16 image patches from it. Then we fix the layer-1 PCA filters, and learn the layer-1 ICA features with the sparse coding algorithm. We expand the dimensions to 512, and update the basis functions for 40000 epochs with a learning rate of 0.1.

We then expanded the receptive fields on the second layer. In each epoch, we randomly sample 64 32x32 image patches, infer the layer-1 ICA feature responses from the four adjacent 16x16 image patches within the 32x32 image patch. All the 512*4=2048 filter responses are fed to the second layer sparse PCA, and the dimensions are reduced to 128. The dimensions are then expanded to 256 by the layer-2 sparse coding algorithm.

## 3    Results

We apply the algorithms to the Kyoto natural scene dataset. The first layer of sparse PCA, as described above, results in center-surround receptive fields (see Figure 1). We have also applied the algorithm to 1) color images, 2) video, and 3) sound. In these cases the results are 1) red-green, blue-yellow, and grayscale opponency, 2) fast, large receptive fields (as in the parasol/magnocellular system), and small, persistent receptive fields (as in the midget/parvocellular system), and 3) gammatone filters. This suggests that our three constraints are somehow implemented by the perceptual system for efficient encoding at the peripheral level.

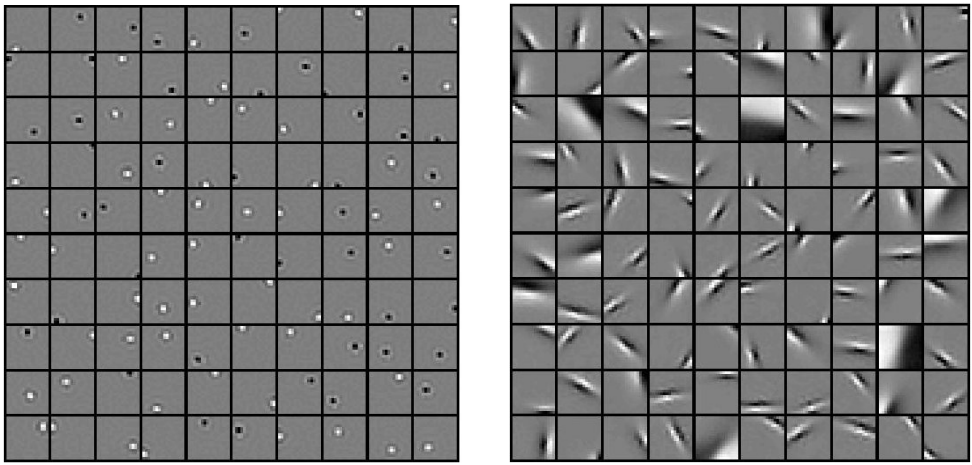

Figure 2. Left: The receptive fields of sparse PCA on natural images. Right: We get the standard result on the output of the sPCA algorithm: oriented local edge detectors.

Applying the sparse coding algorithm to the output of the sPCA, we get the standard result: oriented, local, edge detectors. Now, the interesting thing is what happens after applying sPCA to the output of the first ICA layer. The receptive fields group together oriented edges of the same orientation in a local receptive field. That is, they automatically do a pooling operation, like complex cells in V1 (see Figure 3). Note that a positive connection means that this layer-2 PCA feature prefers strong responses, either positive or negative, from that layer-1 ICA feature. A negative connection means that it prefers weak or no responses from that layer-1 ICA feature. The latter result shows that these model cells also show OFF-pooling, represented by the cold colors.

In order to qualitatively compare our sPCA features to complex cells, we performed a spike-triggered covariance analysis (STC). Neuroscientists have used this to characterize the receptive fields of complex cells in V1. They present animals with white noise stimulation $x$ and record the neuron's response. Then they calculate the average weighted covariance matrix: $C = E[sxx']$. The eigenvectors of $C$ with the biggest eigenvalues are those stimuli that would best activate this neuron; those eigenvectors with smallest eigenvalues are those stimuli that would most suppress this neuron. An example from [Chen et al, 2007] is shown in Figure 4 (left). We plot the STC analysis of a layer-2 sPCA model neuron in the same manner, in Figure 4 (right).

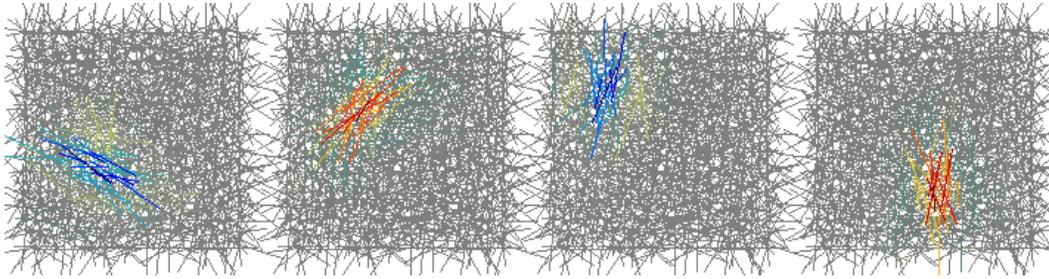

Figure 3. The receptive fields of four randomly-selected units in the second sPCA layer. Within each patch, each bar represents one layer-1 ICA feature. Each layer-1 ICA feature is fitted to a Gabor function, and the location, orientation, and length of the bar represent these features of the Gabor fit. The colors of the bars represent the connection strength from the layer-1 ICA feature to the layer-2 sPCA feature: warm colors, positive connections, cold colors, negative, and gray represents connections near 0.

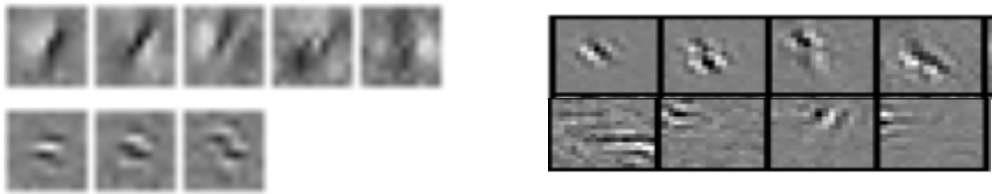

Figure 4. Left: The top row displays the stimuli that would best activate the recorded neuron; the bottom row displays the stimuli that would most suppress this neuron (Chen, Han, Poo, & Dan, PNAS, 2007). Right: The same analysis performed on a layer-2 sPCA model neuron.

There is not a general consensus about what V2 cells represent. We choose a recent paper to compare our model to [6], because they plot the responses of their cells in a similar manner to the technique we use here. They recorded 136 V2 cells from 16 macaque monkeys, but only reported the results on 118 of them. The 18 cells they discarded showed no orientation preference. For each V2 cell, they first identified its classical receptive field. Then they displayed 19 bars arranged in hexagonal arrays within the receptive field, whose sizes were much smaller than the receptive field size. They varied the orientations of the bars, and measured the V2 neurons' responses to those settings. Using this approach, they were able to obtain a space-orientation RF map for each V2 neuron. The result was that about 70% of the 118 neurons showed uniform orientation tuning across their visual field. The remaining cells showed non-uniform tuning (e.g., like the "corner" cell shown in Figure 5, right).

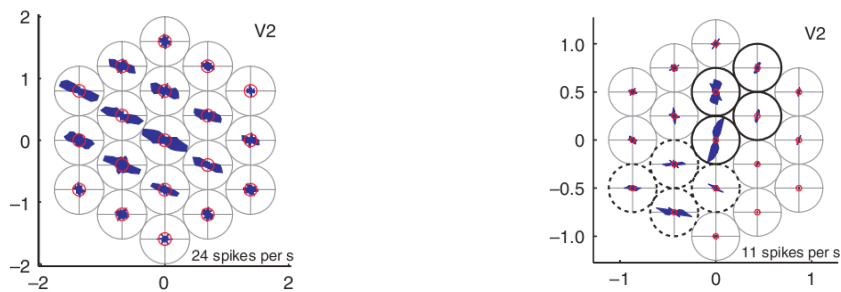

Figure 5 Two plots of the RF maps from the Van Essen experiment. The left plot represents about 70% of the 118 V2 neurons analyzed. These neurons show uniform orientation tuning across their visual field. The right plot represents an example of the remainder of the cells, which show non-uniform orientation tuning.

Six examples of our layer-2 ICA features are plotted in Figure 6, in the same manner as the layer-2 sPCA features in Figure 3. The left-most column displays two model neurons that show uniform orientation preference to layer-1 ICA features. The middle column displays model neurons that have non-uniform/varying orientation preference to layer-1 ICA features. The right column displays two model neurons that have location preference, but no orientation preference, to layer-1

ICA features. Van Essen's group threw out about 13% of the neurons they recorded, because they showed no orientation preference. We suggest that the cells in the right column are a prediction for the receptive fields of the discarded neurons. This preference for location but not shape could support recognition of contours.

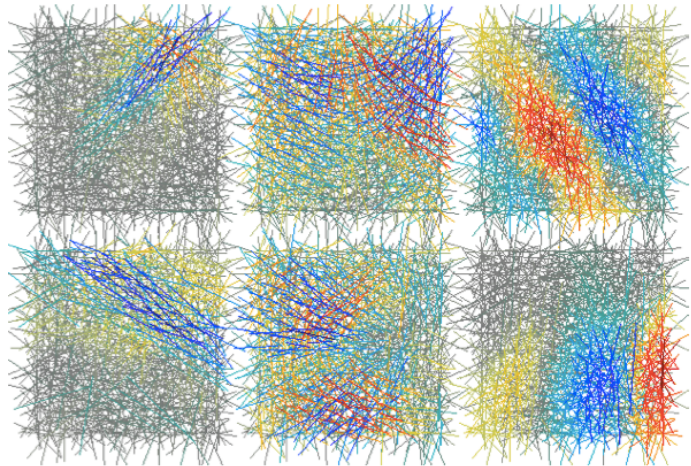

Figure 6. The layer-2 ICA features. The left-most column shows two cells with uniform orientation preference. The middle column shows two cells with non-uniform preference. These two types of cells are roughly in the same proportion as those found by van Essen. About 24% of our cells resemble the ones in the right hand column, which show location preference but no orientation preference.

## 4    Conclusion

We have presented a hierarchical ICA model that is able to capture the characteristics of retinal ganglion cells, V1 simple cells, V1 complex cells, and V2 cells. The model automatically generates representations consistent with the pooling operation of deep networks. It also makes predictions that are consistent with V2 cells that were rejected because of a lack of orientation specificity.

**Acknowledgments**

This work was supported by NSF grants # #IIS-1219252, and SMA-041755 to the Temporal Dynamics of Learning Center, an NSF Science of Learning Center. We would like to thank the members of Gary's Unbelievable Research Unit for helpful comments.